# MIRAD - A comprehensive real-world robust anomaly detection dataset for Mass Individualization

Pulin Li[a], Guocheng Wu[a,*], Li Yin[b], Yuxin Zheng[a], Wei Zhang[c], Yanjie Zhou[a]

[a] School of Management, ZhengZhou University, Zhengzhou, Henan, 450001, China
[b] Department of Data and Systems Engineering, the University of Hong Kong, Hong Kong, China
[c] National Key Laboratory of Equipment State Sensing and Smart Support, College of Intelligence Science and Technology, National University of Defense Technology, Changsha 410073, China
Corresponding author: Guocheng Wu, wu33learn@163.com

**Abstract:**

Social manufacturing leverages community collaboration and scattered resources to realize mass individualization in modern industry. However, this paradigm shift also introduces substantial challenges in quality control, particularly in defect detection. The main difficulties stem from three aspects. First, products often have highly customized configurations. Second, production typically involves fragmented, small-batch orders. Third, imaging environments vary considerably across distributed sites. To overcome the scarcity of real-world datasets and tailored algorithms, we introduce the Mass Individualization Robust Anomaly Detection (MIRAD) dataset. As the first benchmark explicitly designed for anomaly detection in social manufacturing, MIRAD captures three critical dimensions of this domain: (1) diverse individualized products with large intra-class variation, (2) data collected from six geographically dispersed manufacturing nodes, and (3) substantial imaging heterogeneity, including variations in lighting, background, and motion conditions. We then conduct extensive evaluations of state-of-the-art (SOTA) anomaly detection methods on MIRAD, covering one-class, multi-class, and zero-shot approaches. Results show a significant performance drop across all models compared with conventional benchmarks, highlighting the unresolved complexities of defect detection in real-world individualized production. By bridging industrial requirements and academic research, MIRAD provides a realistic foundation for developing robust quality control solutions essential for Industry 5.0. The dataset is publicly available at https://github.com/wu33learn/MIRAD.

**Keywords:** Social manufacturing; Mass individualization; Anomaly detection; Benchmark dataset; Individualized production

# Introduction

Social manufacturing is a service-oriented paradigm that enables communities to self-organize and utilize shared resources to meet customized and long-tail demands (Jiang, 2019). Characterized by fragmented, high-variety, and low-volume orders, social manufacturing networks serve as a key enabler of mass individualization. This pursuit of individualized production represents a central trend in Industry 5.0, which emphasizes placing human needs at the core of manufacturing. The objective is to achieve large-scale production of customized products while maintaining efficiency, sustainability, and creativity. For example, in the footwear industry, consumers can design unique sneakers by choosing from a wide range of options, including color, size, shape, material, and pattern, resulting in products tailored precisely to individual preferences.

While enabling mass individualization, the shift to social manufacturing also presents major hurdles for quality control. Traditional defect detection methods, which depend on stable, high-volume, and standardized outputs, are poorly suited to this new paradigm.

Figure 1 illustrates three interrelated challenges that emerge directly from this context. First, fragmented demand drives dynamic small-batch production, further intensifying the long-tail distribution of defect samples. This creates a dual challenge for supervised learning: critically scarce data for rare defect types and prohibitively high annotation costs. Second, mixed-flow production of individualized goods requires flexible recognition methods that bridge single-class and multi-class identification. Detection systems must cope with constant configurational variations across modules, while simultaneously identifying defects in both standardized and customized components. Moreover, the intricate patterns and diverse colors of individualized designs often closely resemble real flaws, making accurate distinction exceptionally difficult. Third, the geographic distribution and operational flexibility of social manufacturing nodes introduce environmental heterogeneity that undermines detection reliability. Differences in equipment and processes across sites may trigger concept drift, leading to model errors. Equally challenging, variations in imaging environments, such as lighting, camera setups, and background complexity, consistently test model robustness.

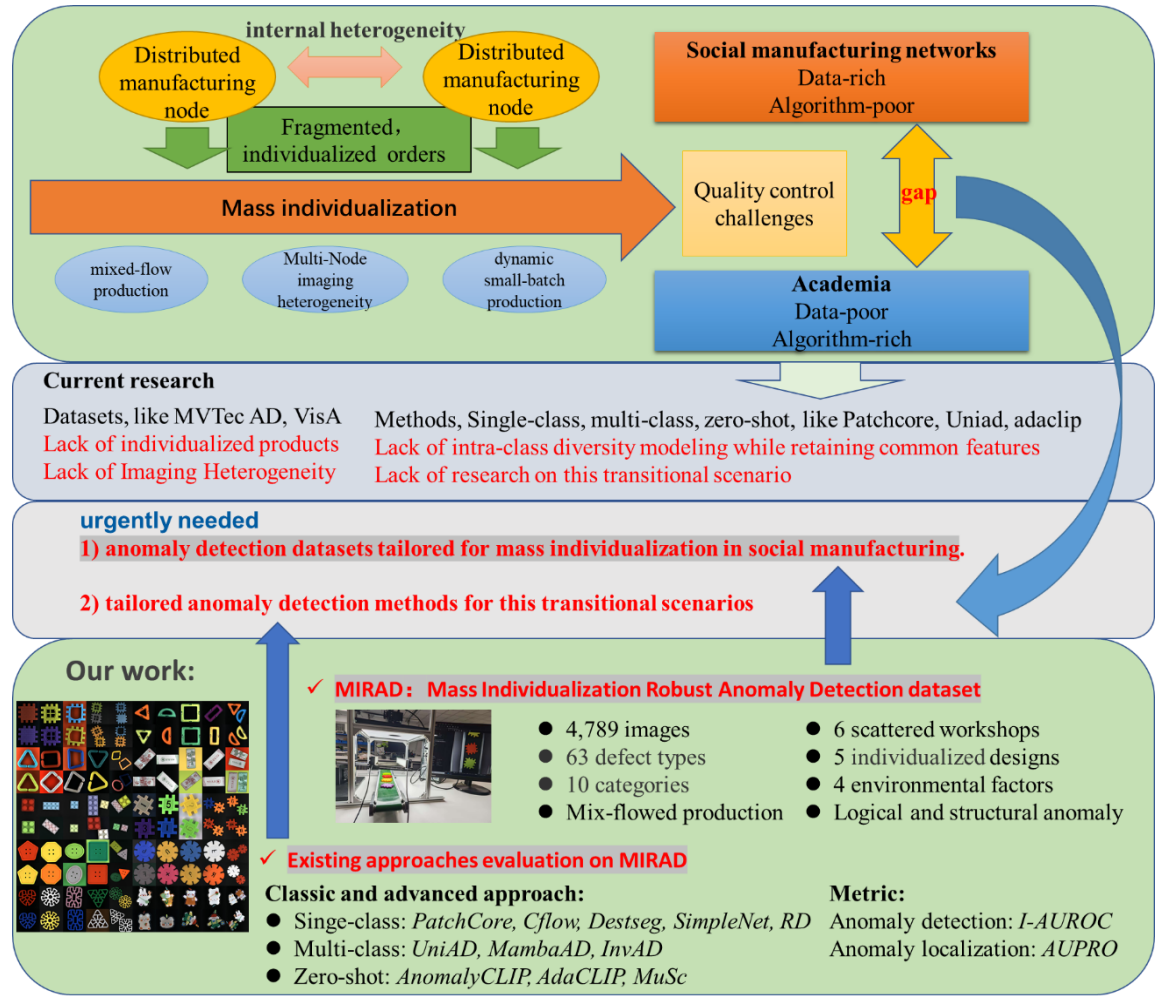

Fig. 1. Challenges of quality control under the social manufacturing paradigm and critical research gaps in anomaly detection.

Individually, each of these three factors challenges defect detection on its own. In combination, they substantially complicate quality control in social manufacturing. Current research on industrial defect detection remains inadequate for this setting, highlighting a fundamental disconnect between industry and academia. On one hand, enterprises hold massive volumes of real-world production data but lack algorithms tuned to these conditions. On the other hand, researchers design advanced models yet struggle to access realistic, complex datasets essential for validation and deployment. This gap, as Fig. 1 underscores, crystallizes into two critical and interdependent research challenges.

(1) **Scarcity of Practical Anomaly Detection Datasets for Social Manufacturing.** Enterprises in social manufacturing networks generate vast volumes of heterogeneous, individualized product data across distributed facilities. Despite this data abundance, publicly available benchmarks capturing real-world complexities remain scarce. Established datasets like MVTec AD and VisA focus on controlled, standardized mass production and cannot represent the operational reality of social manufacturing. Key challenges include extreme product heterogeneity, uncontrolled environments, and fragmented order flows. Most critically, existing anomaly detection methods remain largely unvalidated in mixed-flow individualized production environments. This disconnect creates a fundamental imbalance that enterprises hold rich data, yet the community lacks benchmarks and algorithms tailored to these production streams.

(2) **Scarcity of specialized anomaly detection methods in social manufacturing.** Enterprises operating in social manufacturing environments exhibit high heterogeneity, producing diversified customized products across scattered production units. This heterogeneity demands specialized anomaly detection methods that accommodate product customization and environmental drift. Recent studies have developed one-class, multi-class, and zero-shot models that achieve strong results on standard benchmarks. Yet these models degrade substantially in settings with heterogeneous data sources, product customization, and variable operating conditions. Unpredictable product variation inherent in mass customization blurs the boundary between known and unknown defects, creating a semi-open-set recognition problem. Consequently, a significant disconnect persists between theoretical advancement and practical implementation, as many advanced models remain little tested on the unstructured, complex data found in real production.

To bridge these gaps within social manufacturing, this research introduces two principal contributions to enhance anomaly detection for mass individualization, as illustrated in Fig.1. The first is the development of the Mass Individualization Robust Anomaly Detection dataset, abbreviated MIRAD. MIRAD captures the distributed and dynamic character of social manufacturing networks through the integration of three critical dimensions vital for mass individualization. These dimensions include Mass Individualization itself, characterized by diverse product categories allowing extensive geometric, semantic, and stylistic customization. Further, Multi-node Data Collection gathers products from distributed, heterogeneous manufacturing nodes to realistically capture surface, shape, and logical defects. Finally, Imaging Heterogeneity incorporates images from both standardized and varied visual environments, covering factors like lighting, background, and motion, to accurately mirror real-world production conditions.

The second contribution leverages MIRAD to investigate the performance of established anomaly detection models when processing individualized products originating from distributed social manufacturing networks. This research adopts an unsupervised deep learning paradigm for anomaly detection. This paradigm learns representations solely from normal samples, identifying defects as statistical deviations. Unlike traditional machine vision techniques requiring manual feature engineering, the approach utilizes advanced neural network architectures including Convolutional Neural Networks and Vision Transformers to autonomously extract deep features. It achieves seamless integration of feature extraction and defect detection through end-to-end learning, thereby enhancing adaptability and generalization capabilities. Anomaly detection demonstrates particular efficacy in optimized production lines where defective samples are scarce, contrasting with conventional object detection methods that depend on labeled defect data to learn specific defect characteristics. This method effectively identifies a wide spectrum of defects, encompassing both structural and logical anomalies, but does not provide detailed defect classification.

Therefore, we organize the anomaly detection evaluation on MIRAD across three progressively complex methodological tiers, reflecting escalating demands for model generalization and adaptability. Foundational tier methods encompass representative single-class techniques spanning reconstruction-based, data augmentation-based, and representation-based approaches. Further extending the scope, we incorporate hybrid methods engineered for multi-class anomaly detection, necessitating unified models capable of handling detection tasks across diverse product categories. Finally, the framework integrates zero-shot methods tailored for cold-start scenarios encountered with novel individualized orders. These models typically utilize vision-language models to learn a unified paradigm for identifying anomaly patterns, thereby enabling defect detection without requiring any training samples from the target dataset. This tiered evaluation structure facilitates a comprehensive assessment of SOTA anomaly detection models within mass individualization contexts, rigorously evaluating their generalization capacity and robustness under distributed manufacturing constraints.

The rest of this article is organized as follows. Section 2 reviews the related work on social manufacturing applications, quality control, and existing anomaly detection datasets and methods. In Section 3, a comprehensive description of the proposed MIRAD dataset is provided, detailing the diversity of individualized products and the multi-node imaging heterogeneity. Section 4 establishes the benchmark, including the evaluated methods and evaluation metrics, and presents a thorough analysis of the experimental results for all three types of anomaly detection paradigms. Finally, Section 5 deduces the paper and discusses potential future research directions.

## Related work

This review systematically examines mass customization within the context of social manufacturing, along with relevant quality control strategies. It also summarizes representative anomaly detection datasets and provides an in-depth survey of recent research on one-class, multi-class, and zero-shot anomaly detection approaches.

### Applications and quality control in social manufacturing

**1) Mass individualization applications in industry**

Social manufacturing fundamentally enables mass customization by integrating crowdsourcing, open collaboration, and distributed resources, thereby shifting production paradigms from centralized models to networked and socially-oriented frameworks. This approach leverages community platforms and dispersed assets to underpin mass individualization, facilitating the large-scale production of highly personalized products(Hou et al., 2022). Its application spans diverse industries to achieve individualized production. In footwear, for example, proposed social manufacturing systems utilize 3D scanning, supply-demand matching, and intelligent scheduling to deliver bespoke shoes(Shang et al., 2019). Similarly, lighting brand Gantri democratizes bespoke design through its "Gantri You" initiative, employing 3D printing to offer consumers over 4000 color combinations and fostering user participation via community engagement. Looking ahead, sectors like sportswear point towards significant potential exemplified by initiatives such as Adidas's Speedfactory. This concept envisions localized, small-batch production utilizing robotics, automated knitting systems, and 3D printing, promising product completion within days of order placement and rapid adaptation to regional preferences.

**2) Intelligent quality control in social manufacturing**

The convergence of social manufacturing and customization presents significant challenges to traditional quality control systems, which were designed for batch production. Emerging studies emphasize the need to shift quality assurance upstream by embedding intelligent perception and service-tracking mechanisms throughout the design–manufacturing–delivery lifecycle(J. Liu et al., 2017). For example, cyber-physical systems and IoT platforms can collect process parameters and equipment states in real time, enabling in-process monitoring and early warning.

Intelligent quality management systems, enhanced with big data analytics and machine learning, can model historical defect patterns to predict potential failures, allowing dynamic interventions that reduce defect rates and improve overall efficiency. In additive manufacturing, researchers have developed online quality monitoring systems based on deep convolutional neural networks, achieving over 94% accuracy in identifying and classifying structural and surface defects, with real-time correction capabilities(X. Li et al., 2020). Moreover, in distributed and modular production environments, the integration of multi-agent systems with industrial IoT allows for dynamic task scheduling and adaptive process control, further ensuring consistent product quality(Stefanova-Stoyanova & Stankov, 2020).

## Anomaly detection datasets

Currently, numerous datasets exist for industrial defect detection, which I will review primarily from the perspectives of manufacturing domain and imaging environments. The NEU dataset, one of the earlier contributions, focuses solely on grayscale images of hot-rolled strip steel surfaces, annotating six typical surface defects, including patches (Pa), cracks (Cr), and inclusions (In). In the chip manufacturing sector, Wang et al. (2022) introduced a hybrid mode wafer defect dataset that encompasses eight basic defect types and 29 mixed defect patterns(J. Wang et al., 2020). The dataset that has significantly propelled the advancement of anomaly detection is the MVTec AD dataset. This dataset includes images of five textures and ten objects, featuring over 70 different types of defects such as scratches, dents, contamination, and structural changes (Bergmann et al., 2019). MVTec AD serves as the benchmark dataset for anomaly detection, with SOTA methods achieving an impressive performance score of 0.99. However, some researchers contend that MVTec AD is collected under controlled conditions, where images of the same category are roughly aligned at the pixel level against a clean background, which may not accurately reflect real industrial environments. In contrast, the MPDD dataset captures images of six painted metal parts using varied backgrounds and product rotations (Jezek et al., 2021). Recently, Cheng (2024) introduced the Robust Anomaly Detection dataset, which features free views, uneven illumination, and blurry collections (Cheng et al., 2024).

Additionally, several datasets focus on outdoor maintenance inspections. The MIAD dataset comprises seven outdoor maintenance inspection scenarios and considers uncontrolled outdoor factors such as viewpoints, backgrounds, and surfaces (Bao et al., 2023). Recently, there has been emerging research on defect detection for customized products. Lei (2024) proposed the Texture AD dataset, which addresses various specifications of similar products and includes a diverse array of texture images on surfaces such as cloth, semiconductor wafers, and metal plates (Lei et al., 2024).

Table 1. Comparison of datasets for defect detection. Our MIRAD uniquely captures mass individualization in social manufacturing networks with diverse imaging environments.

| Datasets | category | customized | Defect types | Train | Test | Scene | Sampling Environment |
|---|---|---|---|---|---|---|---|
| NEU-det | 1 | ✗ | 6 | 1,800(total) | | Mass production | monotone |
| MixedWM38 | 1 | ✗ | 8 | 38000(total) | | Mass production | monotone |
| MVTec AD | 15 | ✗ | 73 | 3629 | 1725 | Mass production | monotone |
| MPDD | 6 | ✗ | 11 | 888 | 458 | Mass production | Diverse |
| RAD | 4 | ✗ | / | 213 | 1297 | Mass production | Diverse |
| MIAD | 7 | ✗ | 13 | 70000 | 35000 | Mass production | Diverse |
| Texture AD | 3 | ✔ | / | 28973 | 14147 | Mass customization | monotone |
| MIRAD | 10 | ✔ | 63 | 2398 | 2391 | **Social manufacturing** | Diverse |

Table 1 compares MIRAD with existing datasets. Unlike most benchmarks built on mass production lines with standardized products, MIRAD is collected in social manufacturing networks that support mass individualization. It includes 10 categories of customized products, each varying in color, shape, size, and texture, which better reflects real-world personalization demands. In addition, previous datasets are usually captured in controlled environments with uniform imaging conditions. MIRAD, however, is gathered from six manufacturing nodes across different regions, where illumination, camera settings, working distances, and resolutions vary significantly. This diversity introduces realistic complexity rarely found in earlier datasets. By combining product diversity with environmental heterogeneity, MIRAD offers a more challenging and realistic benchmark. Models trained on MIRAD are therefore more likely to achieve robust and generalizable defect detection across customized products and uncontrolled industrial environments.

**Anomaly detection methods**

Current anomaly detection can be categorized into one-class, multi-class, and zero-shot approaches. While all share the advantage of not needing defect samples for training, each excels in different scenarios. One-class methods specialize in modeling and detecting anomalies within a single known category, often delivering high accuracy for that specific domain. Multi-class approaches are designed to process detection tasks across diverse product categories. Zero-shot techniques offer a unique capability that they can detect anomalies without any training on the target dataset, making them particularly valuable for cold-start situations.

**1) Single-class anomaly detection**

Single-class anomaly detection can be categorized into three types (J. Zhang, He, et al., 2024). Augmentation-based methods introduce defect regions or feature noise into normal images to synthesize abnormal samples. This approach enables the simultaneous learning of features from both normal and abnormal samples and is often referred to as pseudo-supervised learning. Besides, embedding-based approaches leverage pre-trained networks to extract representations of normal features and evaluate anomalies within high-dimensional feature space. Common strategies include maintaining a memory bank of normal features, employing a teacher-student architecture for knowledge distillation, and utilizing normalization flows to map normal image features into a standard normal distribution. Moreover, reconstruction-based techniques utilize an encoder-decoder structure to learn how to reconstruct normal samples. During the detection phase, the inability to effectively reconstruct defective components allows for the localization of defects through the analysis of reconstruction errors.

**2) Multi-class anomaly detection**

Multi-class industrial anomaly detection utilizes a unified model to process detection tasks across diverse product categories. It is adaptive for flexible and individualized production lines, as mixed-flow production of multi category or individualized products creates significant variations in normal and anomalous features. This challenge intensifies in social manufacturing networks, where distributed small-scale production nodes must accommodate multi-class, individualized, and low-volume orders. In such scenarios, traditional single-class scheme suffers from cross category feature interference that may lead to false detection, while also encountering computational and cost inefficiencies in deployment.

Current research in multi-class anomaly detection advances in serval techniques. Firstly, hybrid architectural modeling (UniAD(You et al., 2022), MambaAD (He et al., 2024), ViTAD (J. Zhang, Chen, et al., 2024)) integrates global semantic context with local structural details through multi-scale representation learning, enhancing fine-grained anomaly discrimination. Another important research centers on generative representation disentanglement, employing diffusion-based reconstruction (DiAD (He et al., 2023)) and GAN inversion (InvAD (J. Zhang, Wang, et al., 2024)) to optimize the latent space such that normal and anomalous patterns can be effectively separated. Simultaneously, cross modal constraint strategies (CNC (X. Wang et al., 2025)) have been proposed to align visual features with semantic prototypes, thereby sharpening decision boundaries and suppressing decoder over-generalization on anomalous samples. Further advancements explore dynamic expert systems, incorporating differentiable routing strategies to adaptively allocate computational resources for specialized defect sub-class detection.

### 3) Zero-shot anomaly detection

Zero-shot anomaly detection (ZSAD) refers to the identification of anomalies in images belonging to categories that have not been previously encountered, without the need for any training images from those categories. This approach holds significant promise for addressing cold-start challenges, such as privacy-preserving medical diagnoses and the early stages of production. Current ZSAD models primarily leverage pre-trained visual-language models (VLMs) like CLIP (Radford et al., 2021), DINO (Caron et al., 2021), and SAM, due to their robust generalization capabilities. Some ZSAD methods( WinCLIP (Jeong et al., 2023)) utilize VLMs without any additional training, while other approaches(APRIL GAN (Chen et al., 2023) and AdaCLIP (Cao et al., 2024)) incorporate annotated images from auxiliary anomaly detection datasets to fine-tune VLMs. As long as the test images do not belong to the categories present in the auxiliary dataset, this adaptation aligns with the principles of zero-shot learning. Besides, the auxiliary data may reveal consistent patterns of normal or abnormal instances that VLMs can effectively recognize. On the other hand, several methods apply test data priors to ZSAD. These approaches do not require any training images, while they rely solely on a batch or the complete set of test data for anomaly detection (MuSc (X. Li et al., 2024) and ACR (A. Li et al., 2023)). The fundamental assumption behind is that the occurrence of anomalous images or pixels is typically very low, particularly at the pixel level. Although these methods are labeled as zero-shot, their reliance on test data priors makes them unsuitable for scenarios that require privacy protection.

## MIRAD dataset description

The MIRAD dataset supports social manufacturing nodes to detect defects in mass individualization. It captures the inherent environmental heterogeneity across six distributed manufacturing nodes, where ten individualized products undergo mixed-flow production. Unlike conventional datasets captured in controlled lab environments, MIRAD explicitly embeds four dimensions of node-specific variations, including compound motion patterns, illumination variance, imaging configurations, and background complexity, mirroring the multi-node imaging heterogeneity.

The MIRAD dataset comprises 2398 defect-free training images and 2391 test samples, with 1737 test images containing pixel-level annotated defects. It specifically incorporates multi-object inspection scenarios where two or more individualized products coexist within single frames. The product catalog spans three design paradigms, including geometrically complex components (Geometric Blocks, Buttons), numerically and symbolically encoded items (Number Blocks with Arabic numerals, mathematical symbols), and stylized artifacts (Rabbit Pendants, Magnetic Bookmarks). Both surface defects and logical flaws are captured under node-specific imaging conditions.

## Individualized products diversity

Table 2. Individualized designs and defect of each category in MIRAD

| Category | Individualized designs | | | | | Mixed-flow production | Defect | |
|---|---|---|---|---|---|---|---|---|
| | Color | Shape | Pattern | Size | Semantic | | Structural | Logic |
| Building block | √ | √ | | √ | | √ | √ | √ |
| Chain buckle | √ | √ | | | | √ | √ | √ |
| Geometric block | √ | √ | | √ | | √ | √ | √ |
| Geometric button | √ | √ | | | | √ | √ | √ |
| Geometric snowflake | √ | √ | | | | √ | √ | √ |
| Magnetic block | √ | √ | | | | √ | √ | √ |
| Magnetic bookmark | √ | | √ | | √ | √ | √ | √ |
| Number block | √ | √ | | | √ | √ | √ | √ |
| Number snowflake | √ | | √ | | √ | √ | √ | √ |
| Rabbit pendant | √ | √ | √ | | | √ | √ | √ |

The dataset features a wide range of individualized products that reflect not only the diverse personalization demands from end users, but also the internal heterogeneity of the social manufacturing nodes. Notably, more than 30% of the captured images contain two or more co-occurring individualized products, explicitly demonstrating the mixed-flow production patterns prevalent in social manufacturing. Moreover，the individualized products originate from six distributed production nodes that differ in manufacturing processes, equipment configurations, and image acquisition environments. Specifically, the MIRAD dataset comprises ten categories of personalized products, including *Number Snowflake, Building Block, Chain Buckle, Geometric Block, Magnetic Block, Geometric Button, Magnetic Bookmark, Rabbit Pendant, Geometric Snowflake, and Number Block*. Each category exhibits diverse individualization in geometry, semantics, and styling. The geometric series highlights spatial complexity, while number-themed products incorporate symbolic elements such as Arabic numerals and mathematical notations. Magnetic items vary in both shape and surface texture, and the Rabbit Pendant collection includes ten stylistically unique designs. Other categories, such as Building Block and Chain Buckle, display substantial diversity in form, material, and color. A detailed summary of attribute dimensions is provided in Table 2, with representative examples illustrated in Fig. 2.

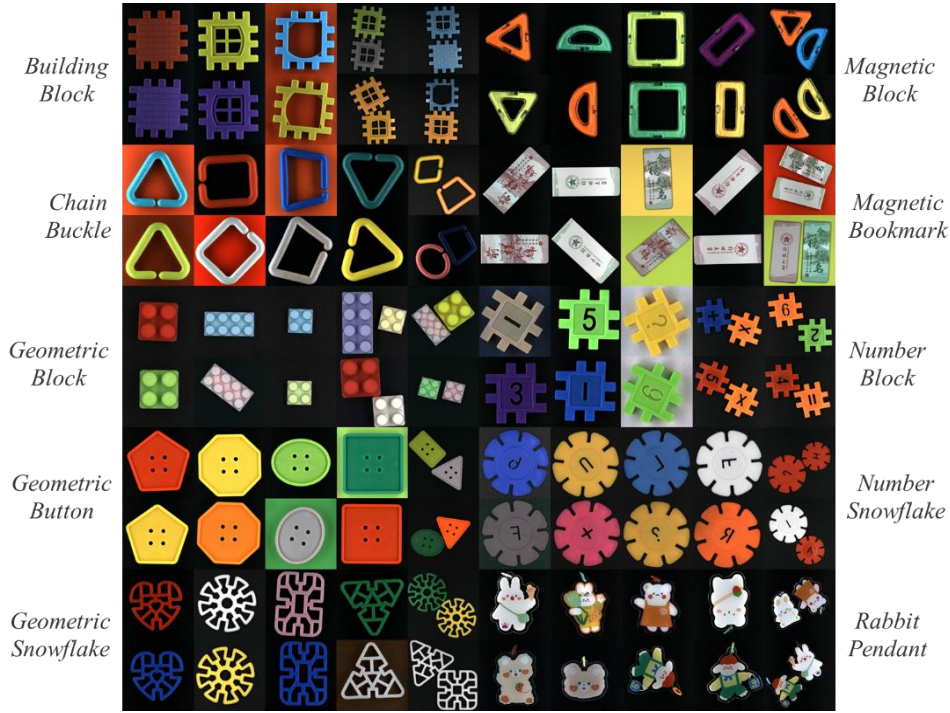

Fig. 2. Individualized products for each category in MIRAD

Defect samples in the test set are derived from real-world production environments and encompass a broad spectrum of types, including both logical and surface-level anomalies. Logical defects are defined as structural inconsistencies resulting from incorrect relative positioning between components, such as missing parts or loose connections between bolts and nuts. Surface defects comprise color inconsistencies, shape deformations, scratches, stains, holes, and other minor visual imperfections. Furthermore, the dataset incorporates complex cases in which multiple defect types co-occur on one or more components within the same instance, thereby increasing the variability and realism of product defects.

For instance, surface defects in the Rabbit Pendant category are observed as scratches, stains, or perforations on the main body, along with color contamination on the hanging rings. The predominant logical defect in this category is the complete absence of the hanging ring. Due to the individualized printed textures and graphical elements, certain minor anomalies are difficult to visually distinguish. One example is the addition of a small red dot on the strawberry-shaped bag element, which can easily be mistaken as part of the original design. Similar subtle defects are observed across other product categories. In total, MIRAD dataset includes 63 manually defined defect patterns, with each product category containing no fewer than six distinct types.

**Multi-Node imaging heterogeneity**

The MIRAD dataset directly incorporates the intrinsic environmental heterogeneity of social manufacturing networks into its data acquisition framework. Since individualized products come from distributed production nodes, they encounter specific workshop imaging conditions that collectively shape social manufacturing quality control. This heterogeneity appears through four key dimensions reflecting real-world production environments. Firstly, illumination setups vary in light type, intensity, and angle primarily based on product material properties, surface textures, and colors. Background settings utilize contrasting color or texture to enhance product visibility. Camera configurations, typically adjusted for object size and motion, differ in exposure mode, resolution, and focal distance. Most products are captured during movement whether translational, rotational, or hybrid which can introduce varying motion blur and positional offsets.

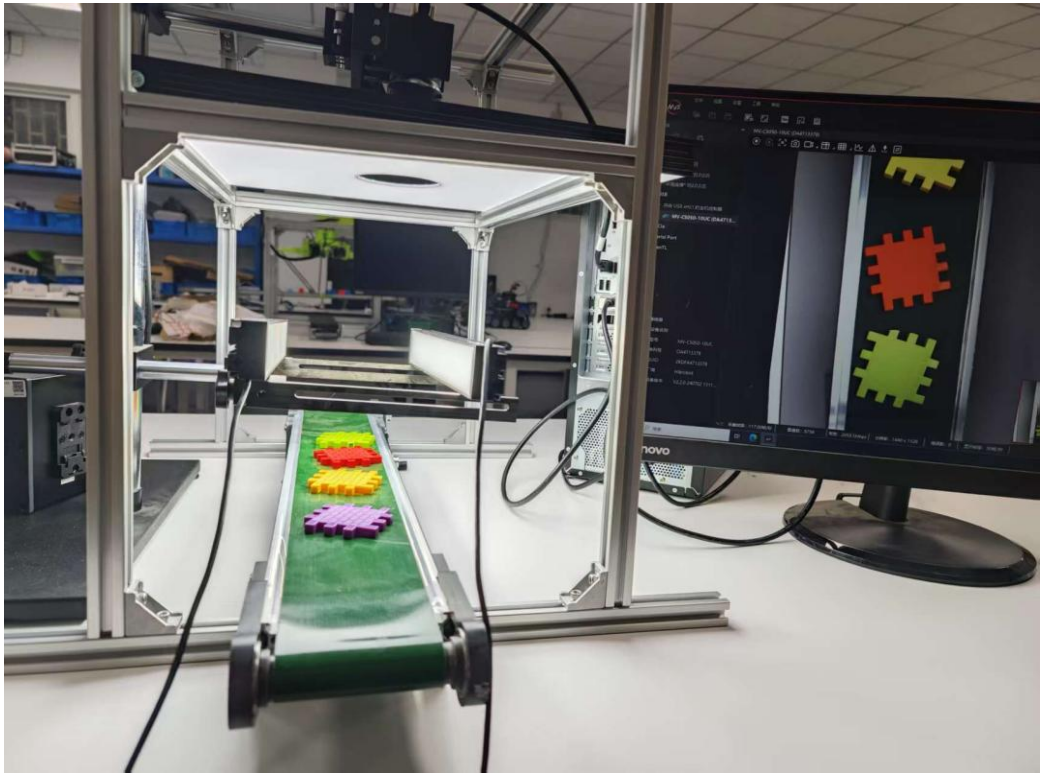

Fig. 3. Individualized rabbit pendant inspection setup in specific social manufacturing node

Fig. 3 shows a production node in the social manufacturing network specializing in custom *Rabbit Pendant* fabrication. These pendants consist of two main parts, including a cartoon-style acrylic pendant and a spray-painted metal hanging ring. This node uses a Hikvision MV-CU200-20GC area-scan camera mounted on a gantry to inspect defects on both custom components. Workers position dual bar lights at an angle to reduce glare while highlighting surface contours. A coaxial LED ring light above provides even illumination. Mid-speed production lines here employ rolling shutter synchronization while high-throughput nodes use global shutter cameras to prevent motion blur.

**1) Compound motion and free view**

MIRAD captures three representative motion types commonly observed in real-world production lines. Linear motion in a conveyor resulting in directional blur, while rotational motion introduces angular variation. Besides, their combination causes position offsets and compound artifacts. Instead of central positioning in laboratory, MIRAD dataset captures objects at varying locations within the frame, reflecting the free-view and compound motion characteristics of flexible mass individualization. Critically, it captures mixed-flow production where multiple individualized products appear together in images. This means MIRAD inherently reflects spatial randomness from flexible routing and visual diversity across distributed workflows. These features provide essential training data for robust quality inspection in real factories.

**2) Diverse illumination conditions**

The varied illumination in social manufacturing networks comes from distributed production workshops customizing their imaging setups. These choices balance local needs like defect detection accuracy, hardware limits, and workflow demands while creating natural visual differences across the network. Unlike standard datasets such as MVTec AD that assume consistent lighting, MIRAD directly shows the distinct lighting choices made by different workshops.

The MIRAD dataset captures real-world illumination variations by acquiring images from six production workshops each with independently configured visual setups. Firstly, forward lighting dominates across nodes as its angular interaction with surfaces effectively highlights structural defects. Dark field illumination at low angles particularly emphasizes surface contours, elevation changes, and topological details, thereby helping identify both concave and convex structures. Besides, exposure parameters vary across nodes, with sampled values including 60,000 μs, 80,000 μs, and 100,000 μs, mirroring actual inspection routines in mass individualization workflows. Moreover, approximately 22% of samples retain ambient light interference where natural sunlight blends with artificial lighting, recreating hybrid illumination conditions typical of open workshop environments.

**3) Diverse resolution conditions**

In mass individualization settings, product sizes can vary greatly because of combined customization choices. Different industrial cameras placed throughout the production network further increase resolution differences. Conventional single resolution training paradigms uses a single resolution, and frequently cannot handle these real-world changes. This is especially true for mixed-flow production lines where individualized items with varying specifications are produced together.

The lens transmission principle underpins multi resolution optical design. This principle states that focal length f divided by working distance d equals image height h divided by object height H. A standardized 60 mm by 60 mm object area H was employed. Three resolutions were implemented using 12 mm industrial cameras 600 by 600, 300 by 300, and 200 by 200 pixels. These configurations yielded calculated working distances of 480 mm, 960 mm, and 1440 mm respectively. This design reflects practical technical requirements. High throughput workshops typically select the shortest 480 mm working distance for rapid inspection cycles. Conversely, precision focused inspection nodes utilize the longest 1440 mm distance to maximize defect identification accuracy.

**4) Rich background**

While earlier datasets like MPDD (Jezek et al., 2021) and MIAD (Bao et al., 2023) provide useful cross domain views, the background settings in MIRAD are carefully designed to mirror the irregularities and variation common in mass individualization workflows. MIRAD includes five distinct background types found across six distributed nodes, each showing the color and texture diversity in real-world production environments.

Background textures and color tones differ across workshops based on their specific operational needs and the products they handle. For example, Workshop B dealing with geometric components, uses ISO-standard neutral gray surfaces. This reduces optical noise during dimensional inspection work. Meanwhile, Workshop D produces decorative items and opts for high contrast orange backgrounds. This choice improves visual clarity significantly, aiding manual quality checks performed there.

```
MIRAD overview
```

The MIRAD dataset bridges the gap between laboratory research and practical quality control in mass individualization under the context of social manufacturing. MIRAD, comprising 10 categories of individualized products, encompasses three design paradigms, including geometrically complex components, numerically and symbolically encoded items, and stylized artifacts. Each product category offers diverse options in color, shape, pattern, and size, covering one or even all three design paradigms. Notably, 30% images of the dataset contain co-occurring customized products, reflecting mixed-flow manufacturing patterns. Central to the dataset's design is its explicit modeling of multi-node environmental heterogeneity inherent to social manufacturing, with data collected from six workshops engaged in mass individualization. Distributed production nodes employ diverse motion and illumination strategies, different camera specifications , and node-specific background, collectively mirroring the diversity of real-world dis manufacturing networks. Further information of each category is summarized in Table 3.

Table 3. Statistical overview of the MIRAD dataset for each category, including the quantity of images in both the training and test sets, the number of defective types, and details about the multi-node imaging heterogeneity of each category. (DV, DI, DR, DB stand for diverse view, illumination, resolution and background, respectively ).

| Category | Train | Test | | Defect types | Imaging Heterogeneity | | | |
|---|---|---|---|---|---|---|---|---|
| | | Normal | Abnormal | | DV | DI | DR | DB |
| Building block | 262 | 57 | 158 | 7 | √ | | √ | √ |
| Chain buckle | 197 | 81 | 141 | 6 | √ | √ | √ | √ |
| Geometric block | 238 | 55 | 159 | 6 | √ | √ | √ | |
| Geometric button | 290 | 94 | 188 | 7 | √ | √ | √ | √ |
| Geometric snowflake | 215 | 62 | 163 | 7 | √ | √ | √ | √ |
| Magnetic block | 268 | 57 | 204 | 6 | √ | | √ | |
| Magnetic bookmark | 216 | 48 | 133 | 1 | √ | √ | √ | √ |
| Number block | 254 | 69 | 151 | 8 | √ | √ | √ | √ |
| Number snowflake | 230 | 60 | 236 | 8 | √ | √ | √ | √ |
| Rabbit pendant | 228 | 71 | 204 | 7 | √ | √ | √ | √ |
| Total | 2398 | 654 | 1737 | 63 | | | | |

Our MIRAD dataset features the following characteristics.

1) Diverse individualized products. The dataset includes objects and fabrics produced in mass customization, highlighting the diversity and differences between samples. These individualized items come in many unique designs featuring variations in color, pattern, material, shape and size.

2) complex and varied imaging environment across nodes. MIRAD embeds environmental heterogeneity inherent to distributed production networks through four dimensions, including compound motion patterns, illumination variance, imaging configurations, and background complexity. Such heterogeneity challenges detectors to decouple contextual noise from defect sensitive features, testing the robustness and generalization of detection models.

3) Data sourced directly from production lines. All data were sourced from six distributed nodes involved in mass individualization. More than thirty percent of captured images contain two or more co-occurring individualized products within single frame, reflecting the mixed-flow production patterns prevalent in social manufacturing. Notably, thirty eight percent of defective samples exhibit compound anomalies where multiple defect types coexist.

## Benchmark

Based on MIRAD, we explore how existing anomaly detection models perform when confronted with the individualized products, as well as their robustness to imaging heterogeneity across distributed production nodes. Specially, we evaluate three kinds of approaches on MIRAD dataset, including single-class, multi-class and zero-shot detection models, providing a foundation for future analyses.

## Evaluated methods

We adopt the publicly available code in Ader (J. Zhang, He, et al., 2024) for both single-class and multi-class approaches. All training and testing images are uniformly resized to 256 × 256 pixels. No data augmentation is applied, and all experiments on the MIRAD dataset are conducted over 100 epochs using a single NVIDIA 4090 GPU.

For zero-shot methods, we utilize the publicly available implementations released by the original authors. Specifically, AdaClip employs pre-trained weights derived from multiple auxiliary datasets, including MVTec AD (Bergmann et al., 2019), VisA(Zou et al., 2022), and ClinicDB (Bernal et al., 2015). In contrast, AnomalyCLIP (Zhou et al., 2024) is initialized with weights pre-trained solely on the VisA dataset.

### 1) Single-class anomaly detection

Classical anomaly detection learns representations exclusively from normal samples and identifies defects as outliers, serving as unsupervised leaning. These methods follow the principle of one-class novelty detection, where a separate model is trained for each product class. We select representative methods from three technical paradigms, including augmentation-based approaches (Destseg (X. Zhang et al., 2023) and SimpleNet (Z. Liu et al., 2023)), embedding-based methods ( PatchCore (Roth et al., 2022) and CFlow (Lee et al., 2022)), and reconstruction-based frameworks (RD (Deng & Li, 2022)), covering key strategies for anomaly detection in industrial scenarios.

**PatchCore** (Roth et al., 2022)**:** PatchCore employs a highly representative memory bank of nominal patch features with an outlier detection model.

**Cflow** (Lee et al., 2022)**:** CFLOW leverages conditional normalizing flows to model the feature distribution of normal samples. By computing exact likelihoods in a multi-scale manner, it enables accurate pixel-level localization of anomalies with low computational overhead.

**Destseg** (X. Zhang et al., 2023)**:** DeSTSeg introduces a segmentation-guided denoising student–teacher framework for unsupervised visual anomaly detection. The student network is trained on synthetically corrupted images, while a segmentation module fuses multi-level feature discrepancies to improve anomaly localization and mitigate over-generalization.

**SimpleNet** (Z. Liu et al., 2023)**:** SimpleNet incorporates Gaussian noise into the extracted normal features to generate abnormal samples. Both normal and corresponding abnormal features are input simultaneously to train the final discriminator.

**RD** (Deng & Li, 2022)**:** RD introduces a reverse distillation framework for anomaly detection, where a student network reconstructs multi-scale features from low-dimensional embeddings produced by a pre-trained teacher. The approach emphasizes feature discrepancy between teacher and student to capture subtle deviations indicative of anomalies.

### 2) Multi-class anomaly detection

Multi-class anomaly detection enables unified models to handle detection tasks across diverse product categories within a single framework, eliminating the need for training and maintaining multiple class-specific detectors. This paradigm becomes particularly essential within flexible or customized production lines, where mixed-flow production of multi-category products creates significant variations in normal and abnormal features. In this scenarios, traditional one-model-per-class scheme suffers from cross-category feature interference that leads to false detection, while also encountering computational and cost inefficiencies in deployment.

Our MIRAD dataset represents a transition setting between single-class and multi-class anomaly detection, where customized products across categories share common features yet retain individualized variations, challenging traditional one-class models and motivating unified multi-class approaches.

**UniAD** (You et al., 2022)**:** UniAD first formally introduce multi-class anomaly detection, aiming to detect anomalies across diverse categories using a single unified model. It introduces a layer-wise query decoder, neighbor-masked attention, and feature jittering to prevent trivial reconstructions and enhance representation diversity, enabling effective modeling of heterogeneous class distributions within a single architecture.

**MambaAD** (He et al., 2024)**:** MambaAD pioneers the application of Mamba architecture to multi-class anomaly detection. It leverages state space models to efficiently capture both global and local dependencies. A locality enhancement module with hybrid scanning mechanisms is employed to reconstruct multi-scale features.

**InvAD** (J. Zhang, Wang, et al., 2024)**:** InvAD adapts GAN inversion for multi-class anomaly detection. The method incorporates a spatial style modulation module for input-dependent reconstruction. This approach significantly enhances detection capability for diverse anomaly types.

**3) Zero-shot anomaly detection**

Zero-shot anomaly detection targets the identification of anomalies from unseen categories, without relying on any class-specific training data. It leverages the generalization capabilities of pre-trained vision-language models to enable anomaly detection in data-sparse scenarios such as early-stage production or sensitive privacy domains.

**AnomalyCLIP** (Zhou et al., 2024)**:** AnomalyCLIP introduces an object-agnostic prompt learning, adapting the CLIP for zero-shot anomaly detection. It learns generalizable textual prompts that represent normality and abnormality independently of specific object categories, enabling detection across diverse domains.

**AdaCLIP** (Cao et al., 2024)**:** AdaCLIP utilizes annotated images from auxiliary anomaly detection datasets to fine-tune the pre-trained CLIP model for anomaly detection tasks. It also develops multimodal hybrid learnable prompts to maximize the utility of auxiliary anomaly detection data.

**MuSc** (X. Li et al., 2024)**:** MuSc notes that for industrial product images, normal image patches can find a relatively large number of similar patches in other unlabeled images, whereas abnormal patches have only a few similar counterparts. It proposes a mutual scoring mechanism to utilize unlabeled test images for assigning anomaly scores to each other.

## Evaluation metric

For anomaly detection, we use the common AUROC metric, which stands for Area Under the Receiver Operating Characteristic curve. This score effectively measures both the True Positive Rate and the False Positive Rate. The I-AUROC metric mainly tests how well a model can distinguish anomalies, working reliably even when the data has significant class imbalances common in industrial settings.

Regarding anomaly localization, pixel-wise decision is adopted to classify individual pixels as normal or anomalous. However, the MIRAD dataset contains very small flaws, allowing models that simply predict every pixel as normal to still get misleadingly high pixel AUROC scores. To overcome this issue, we use the Area Under the Per Region Overlap curve (AUPRO) (Roth et al., 2022) as a more trustworthy metric for detailed defect location assessment. AUPRO equally values anomaly areas of all sizes and calculates the proportional overlap between predicted defect maps and actual ground truth regions. This provides a fairer and more informative evaluation of localization

## Result analysis

**1) Single-class and multi-class approaches**

Table 4 presents performance results for all evaluated methods across every product category in MIRAD. It includes the overall average performance and provides benchmark comparisons using MVTec AD and VisA datasets. These comparisons help illustrate how MIRAD presents greater challenges relative to these established datasets.

(1) Performance on the MIRAD dataset reveals notable limitations. Single-class methods achieved only 76.2% I-AUROC, whereas multi-class approaches scored an even lower score of 71.1%. In contrast, both single-class and multi-class frameworks reached approximately 98% I-AUROC on MVTec AD and around 93% on VisA. This significant performance drop highlights the increased difficulty of defect detection in social manufacturing settings.

(2) Among all evaluated models, Reverse Distillation (RD) achieved the best performance in both defect detection and localization, with I-AUROC and PRO scores of 79.8% and 87.7%, respectively. This suggests that reverse distillation from one-class embeddings may be particularly effective in mass individualization scenarios, since class boundaries are less clearly defined and variations within single category are substantial.

(3) The *Rabbit Pendant* category proved exceptionally challenging for defect detection models. Both one-class and multi-class models achieved only 55% I-AUROC. The *Rabbit Pendant* features diverse cartoon-style acrylic bodies paired with varied spray-painted metal hanging rings. This worsen performance illustrates the difficulty models face in distinguishing defective from normal instances when product designs are highly individualized.

(4) MIRAD represents a hybrid scenario bridging single class and multi class anomaly detection paradigms. Each product in MIRAD dataset is individualized, containing both distinctive individualized characteristics and common shared features. However, no evaluated method, either single-class or multi-class, surpassed 80% I-

AUROC, indicating that tailored approaches are needed to handle such transitional scenario. Notably, single-class methods consistently outperformed multi-class methods on several categories, such as *Geometric Block, Geometric Button, and Magnetic Bookmark*, with I-AUROC improvements of up to 10 percentage points. This suggests that intra-class consistency may offer a stronger modeling advantage than inter-class differences under the MIRAD dataset.

Table 4. performance comparisons on MIRAD and some existing anomaly detection datasets, MVTec AD and VisA.

|  | patchcore | cflow | destseg | simplenet | rd | Mean±Std | uniad | mambaad | invad | Mean±Std |
|---|---|---|---|---|---|---|---|---|---|---|
| Building block | 83.0 | 86.5 | 86.1 | 88.0 | 89.6 | 86.6±2.2 | 72.1 | 85.3 | 87.1 | 81.5±6.7 |
|  | 69.8 | 78.1 | 73.8 | 62.7 | 87.7 | 74.4±8.4 | 65.8 | 79.2 | 83.7 | 76.2±7.6 |
| Chain buckle | 79.7 | 76.2 | 69.2 | 83.7 | 86.3 | 79.0±6.0 | 71.1 | 83.2 | 81.1 | 78.5±5.3 |
|  | 77.1 | 82.6 | 77.3 | 77.2 | 85.6 | 80.0±3.5 | 68.0 | 83.9 | 83.0 | 78.3±7.3 |
| Geometric block | 86.6 | 94.2 | 94.0 | 93.3 | 93.9 | 92.4±2.9 | 62.6 | 88.7 | 93.8 | 81.7±13.7 |
|  | 90.3 | 92.1 | 88.6 | 87.3 | 94.2 | 90.5±2.5 | 75.1 | 93.4 | 93.1 | 87.2±8.6 |
| Geometric button | 77.4 | 86.2 | 76.9 | 92.9 | 88.8 | 84.4±6.3 | 55.6 | 85.3 | 85.0 | 75.3±13.9 |
|  | 81.8 | 84.3 | 90.3 | 80.1 | 96.1 | 86.5±5.9 | 64.3 | 96.5 | 93.2 | 84.7±14.5 |
| Geometric snowflake | 65.8 | 71.4 | 83.3 | 77.4 | 81.0 | 75.8±6.4 | 71.7 | 70.4 | 78.7 | 73.6±3.6 |
|  | 50.0 | 60.3 | 85.6 | 61.1 | 79.8 | 67.4±13.3 | 31.2 | 67.2 | 72.4 | 56.9±18.3 |
| Magnetic block | 61.3 | 64.3 | 60.8 | 69.3 | 73.2 | 65.8±4.8 | 53.7 | 71.4 | 68.5 | 64.5±7.8 |
|  | 77.1 | 82.3 | 77.3 | 63.0 | 90.8 | 78.1±9.0 | 56.6 | 88.1 | 89.7 | 78.1±15.2 |
| Magnetic bookmark | 67.4 | 70.3 | 73.6 | 76.3 | 73.5 | 72.2±3.1 | 47.5 | 59.4 | 68.6 | 58.5±8.6 |
|  | 68.0 | 72.5 | 63.0 | 67.3 | 82.3 | 70.6±6.6 | 54.1 | 73.5 | 76.2 | 67.9±9.8 |
| Number block | 57.3 | 67.3 | 70.8 | 69.3 | 68.9 | 66.7±4.8 | 54.8 | 62.5 | 71.1 | 62.8±6.7 |
|  | 73.2 | 83.5 | 79.5 | 66.2 | 89.0 | 78.3±7.9 | 64.7 | 81.2 | 87.1 | 77.7±9.5 |
| Number snowflake | 77.3 | 81.3 | 85.3 | 82.6 | 86.2 | 82.5±3.2 | 65.6 | 84.3 | 86.3 | 78.7±9.3 |
|  | 86.3 | 81.1 | 83.5 | 75.6 | 93.5 | 84.0±5.9 | 63.0 | 91.1 | 90.8 | 81.6±13.2 |
| Rabbit pendant | 54.9 | 52 | 56.1 | 62.1 | 56.1 | 56.2±3.3 | 54.8 | 57.7 | 54.3 | 55.6±1.5 |
|  | 70.1 | 78.4 | 54.7 | 574 | 78.3 | 67.8±10.1 | 61.7 | 74.3 | 78.9 | 71.6±7.3 |
| MIRAD | 71.1 | 75.0 | 75.6 | 79.5 | 79.8 | 76.2±4.3 | 61.0 | 74.8 | 77.5 | 71.1±7.7 |
|  | 74.4 | 79.5 | 77.4 | 68.8 | 87.7 | 77.8±7.3 | 60.5 | 82.8 | 84.8 | 76.0±11.1 |
| Mvtec ad | 99.0 | 97.1 | 98.4 | 99.3 | 98.5 | 98.5±0.8 | 97.5 | 98.6 | 98.2 | 98.1±0.5 |
|  | 94.7 | 92.3 | 94.7 | 89.9 | 93.1 | 92.9±1.8 | 90.7 | 93.1 | 94.1 | 92.6±1.4 |
| Visa | 95.1 | 91.8 | 90.3 | 95.4 | 95.9 | 93.7±2.2 | 88.8 | 94.3 | 95.5 | 92.9±2.9 |
|  | 91.2 | 84.8 | 90.6 | 88.7 | 92.9 | 89.6±2.8 | 85.5 | 91.0 | 92.5 | 89.7±3.0 |

2) **Zero-shot anomaly detection**

As illustrated in Table 5, the three selected zero-shot models all achieved I-AUROC scores below 70% on the MIRAD dataset, with an average AUPRO of 74.0%. Compared to methods trained directly on MIRAD, both detection and localization performance were consistently lower. Specifically, the *Geometric Snowflake* exhibits the poorest performance with an I-AUROC of 46.2%, followed closely by Rabbit Pendant at 52.6%, indicating the particular difficulty of detecting anomalies in highly individualized products.

Although AnomalyCLIP (Zhou et al., 2024) and AdaCLIP (Cao et al., 2024) have been pre-trained on auxiliary datasets such as MVTec AD(Bergmann et al., 2019) and VisA (Zou et al., 2022), they still fell short on the MIRAD benchmark. The unified representations of normality and abnormality learned from conventional datasets may fail to generalize to MIRAD, where individualized normal features vary widely across instances. When defects co-occur or blend with individualized designs, MIRAD dataset exhibits distinct patterns that challenge models trained on standardized examples.

Similarly, MUSC (X. Li et al., 2024), which leverages prior knowledge from test batches, also performed poorly with an average I-AUROC of 64.2%. This is likely due to the limited diversity of a single test batch, which cannot fully represent the broad range of individualization present in MIRAD.

Table 5. Zero-shot anomaly detection performance on MIRAD, including I-AUROC and AUPRO results across product categories

|  | anomalyclip | adaclip | musc | Mean±Std |
|---|---|---|---|---|
| Building block | 81.9 | 78.1 | 69.4 | 76.5±5.2 |
|  | 58.7 | 65.3 | 76.0 | 66.7±7.1 |
| Chain buckle | 57.1 | 59.6 | 65.2 | 60.6±3.4 |
|  | 76.1 | 78.9 | 81.0 | 78.7±2.0 |
| Geometric block | 86.8 | 87.2 | 83.6 | 85.9±1.6 |
|  | 83.5 | 85.3 | 89.6 | 86.1±2.6 |
| Geometric button | 63.2 | 71.6 | 67.2 | 67.3±3.4 |
|  | 66.8 | 81.6 | 79.3 | 75.9±6.5 |
| Geometric snowflake | 59.4 | 59.4 | 46.2 | 55.0±6.2 |
|  | 41.1 | 65.7 | 64.6 | 57.1±11.3 |
| Magnetic block | 61.7 | 71.3 | 64.1 | 65.7±4.1 |
|  | 70.6 | 83.3 | 81.4 | 78.4±5.6 |
| Magnetic bookmark | 74.4 | 53.7 | 65.0 | 64.3±8.5 |
|  | 70.4 | 71.0 | 81.5 | 74.3±5.1 |
| Number block | 76.0 | 76.5 | 61.5 | 71.3±7.0 |
|  | 67.6 | 77.2 | 83.5 | 76.1±6.5 |
| Number snowflake | 74.4 | 81.7 | 66.8 | 74.3±6.1 |
|  | 77.0 | 84.7 | 83.6 | 81.8±3.4 |
| Rabbit pendant | 56.4 | 34.2 | 52.6 | 47.7±9.7 |
|  | 65.6 | 57.2 | 72.2 | 65.0±6.1 |
| Average | 69.1 | 67.3 | 64.2 | 66.9±2.0 |
|  | 67.8 | 75.0 | 79.3 | 74.0±4.7 |

## Conclusion

To address the quality control challenges in social manufacturing, we introduce the MIRAD dataset, which emphasizes individualized product and multi-node imaging heterogeneity. MIRAD comprises 10 categories of individualized products, each exhibiting diverse characteristics in color, material, shape, size and pattern, filling a critical gap left by existing manufacturing datasets that often lack intra-class variability. We argue that such intra-class diversity constitutes a transitional anomaly detection scenario, situated between traditional one-class and multi-class settings. The dataset was collected from six distributed production nodes under compound motion, free-view imaging, dynamic illumination, variable resolution, and background complexity, thereby capturing the multi-node imaging heterogeneity intrinsic to social manufacturing. These conditions pose new challenges for the robustness and generalizability of current anomaly detection models.

MIRAD mirrors the distributed, dynamic nature of social manufacturing networks, as well as integrating three critical dimensions in mass individualization scenarios. (1) Mass individualization. MIRAD includes 10 categories of individualized products, each exhibiting diverse customization in geometry, semantics, and styling. Some products feature intricate geometric shapes, such as *Geometric Block*, *Button*, and

*Snowflake*. Others adopt numerical or symbolic designs, such as *Number Block* and *Snowflake*, which incorporate Arabic numerals, mathematical symbols, and English words. The remaining categories include uniquely styled items such as the *Rabbit Pendant* and magnetic products like the *Magnetic Block* and *Bookmark*. (2) Multi-node data collection in social manufacturing. Individualized products are derived from six production nodes, directly reflecting the socialized, distributed, and heterogeneous manufacturing resources. This multi-node sourcing strategy enables comprehensive coverage of real-world defect patterns, including surface imperfections (e.g., color variations and scratches), shape deformations, as well as logical flaws such as component misalignments. (3) Imaging heterogeneity across production nodes. To capture the environmental heterogeneity inherent in distributed production, MIRAD includes not only images acquired under standardized conditions but also samples collected under diverse visual settings. These include variations in lighting, background color, and object pixel density, reflecting differences across production nodes. Moreover, products are recorded under motion scenarios to simulate the dynamic and often unstable acquisition conditions encountered in real-world distributed workshops.

We conducted thorough evaluation of SOTA single-class, multi-class, and zero-shot anomaly detection methods using the MIRAD benchmark. Despite their established effectiveness on standard datasets like MVTec AD and VisA, all approaches showed striking performance gaps when applied to MIRAD. This outcome validates the substantially greater challenge of identifying defects in individualized products within distributed production environments. Zero-shot approaches pretrained on auxiliary datasets, including MVTec AD and VisA, still performed worse than models directly trained on MIRAD. These results demonstrate that the unified representations of normality and abnormality learned from conventional datasets do not transfer well to individualized production scenarios. Consequently, MIRAD emerges as a valuable counterpart to traditional mass-production benchmarks, more accurately reflecting the complexities of decentralized, individualized manufacturing.

We hope that MIRAD will encourage the research community to direct greater attention toward defect detection in mass individualization, advancing robust quality assurance solutions for social manufacturing environments.